\documentclass[letterpaper, 10 pt, conference]{ieeeconf}  

\IEEEoverridecommandlockouts                              

\usepackage{graphics} 
\usepackage{epsfig} 
\usepackage{mathptmx} 
\usepackage{times} 
\usepackage{amsmath} 
\usepackage{amssymb}  
\usepackage{booktabs}
\usepackage{algorithm}
\usepackage{algorithmicx}
\usepackage{algpseudocode}
\usepackage{indentfirst}
\usepackage{multirow}
\usepackage{hyperref}
\usepackage{marvosym}
\usepackage{color}

\title{\LARGE \bf
Evaluation of Pedestrian Safety in a High-Fidelity Simulation Environment Framework}

\author{Lin Ma$^{1}$*, Longrui Chen$^{1}$*, Yan Zhang$^{1}$*, Mengdi Chu$^{1}$, Wenjie Jiang$^{1}$, Jiahao Shen$^{1}$,Chuxuan Li$^{1}$, \\ Yifeng Shi$^{2}$, Nairui Luo$^{2}$, Jirui Yuan$^{1}$, Guyue Zhou$^{1}$, and Jiangtao Gong$^{1}$\textsuperscript{\Letter}
\thanks{*The three authors contribute equally to this work.}
\thanks{$^{1}$Institute for AI Industry Research (AIR), Tsinghua University, 10080, Haidian District, Beijing, P.R.China.
        {\tt\small secondnamefirstname@air.tsinghua.edu.cn}}%
\thanks{$^{2}$Baidu Inc.
}%
}

\begin{document}

\maketitle
\thispagestyle{empty}
\pagestyle{empty}

\begin{abstract}

Pedestrians' safety is a crucial factor in assessing autonomous driving scenarios. However, pedestrian safety evaluation is rarely considered by existing autonomous driving simulation platforms.
This paper proposes a pedestrian safety evaluation method for autonomous driving, in which not only the collision events but also the conflict events together with the characteristics of pedestrians are fully considered. Moreover, to apply the pedestrian safety evaluation system, we construct a high-fidelity simulation framework embedded with pedestrian safety-critical characteristics. We demonstrate our simulation framework and pedestrian safety evaluation with a comparative experiment with two kinds of autonomous driving perception algorithms---single-vehicle perception and vehicle-to-infrastructure (V2I) cooperative perception.
The results show that our framework can evaluate different autonomous driving algorithms with detailed and quantitative pedestrian safety indexes. To this end, the proposed simulation method and framework can be used to access different autonomous driving algorithms and evaluate pedestrians' safety performance in future autonomous driving simulations, which can inspire more pedestrian-friendly autonomous driving algorithms.

\end{abstract}

\section{Introduction}
 Safety is a critical issue for autonomous driving, which must be tested in a simulation environment before it is tested on real-world roads. Pedestrian safety is an important factor in assessing autonomous driving scenarios because pedestrians account for more than one-fifth of all fatalities in road accidents worldwide, dating back to the first reported pedestrian fatality in 1899~\cite{world2015global}.

The collision rate is the most intuitive but one-sided evaluation for vehicle safety performance. As a supplement, collision severity is also commonly used. When evaluating pedestrian safety, both collision rate and collision severity should be taken into account, as mentioned in ~\cite{SHEYKHFARD2021681}. However, the evaluation of pedestrian safety is more complicated than that of vehicle safety for the following reasons:
\begin{itemize}
\item Non-collision scenarios are equally important for pedestrian safety. Conflict scenarios often occur in real driving environments~\cite{su15010858}. While these scenarios do not directly result in a collision that causes injury to pedestrians, they can cause panic among pedestrians, which is also a critical factor for pedestrian safety.
\item Pedestrians with different characteristics react differently even in the same dangerous scene, so the collision rate varies depending on these factors ~\cite{LIU2021115}. For instance, young adults tend to respond significantly faster to emergency situations than children and the elderly.
\item Pedestrians with different characteristics, particularly those of different ages, experience varying levels of injury severity in the same type of collision. For instance, the severity of injury increases with age at the same collision intensity ~\cite{DELEN2006434}.
\end{itemize}

Thus, when evaluating pedestrian safety, it is crucial to consider all possible scenarios and pedestrian characteristics thoroughly.

\begin{figure}
    \centering
    \includegraphics[scale=1.3]{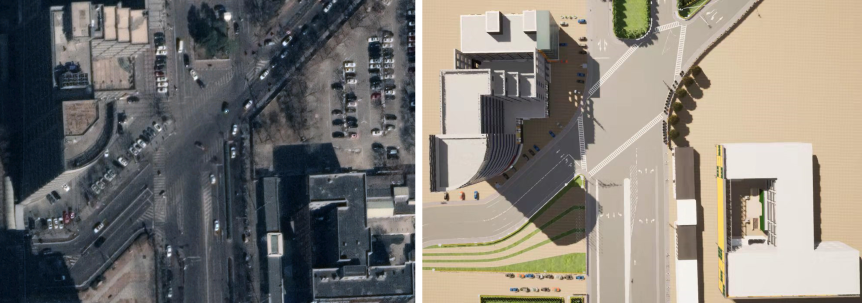}
    \caption{The real-world and simulated scenario of an urban intersection in the high-fidelity simulation framework.}
    \label{fig:digitaltwin}
\end{figure}

Before testing on real roads, the safety performance of autonomous vehicles, as well as pedestrian safety, needs to be verified in a simulation environment. However, embedding the pedestrian safety evaluation module in a high-fidelity simulation framework remains challenging. Most commonly used simulation frameworks are based on numerical or traffic simulation, such as ProVerif~\cite{ta2020secure}, PLEXE~\cite{jia2016enhanced}, MATLAB~\cite{mughal2020cooperative}, SUMO~\cite{he2021improved}, and VISSIM~\cite{yan2018coordinated}. These frameworks lack a quantitative evaluation model that comprehensively considers all safety-critical scenarios and pedestrian characteristics~\cite{kalra2016driving}, leading to biased evaluations of pedestrian safety.

To address this issue, this paper presents a high-fidelity simulation environment framework that constructs a digital twin of an urban intersection in China (Fig. \ref{fig:digitaltwin}). 
In addition to highly reductive the road structure, this framework also takes the traffic modelling and pedestrian simulation into consideration, which allows for a fair and unbiased evaluation of pedestrian safety in autonomous driving scenarios.

The main contribution of this paper can be summarized as:

\begin{enumerate}
    \item A comprehensive evaluation system for pedestrian safety based on both collision and non-collision scenarios integrated with pedestrian characteristics. 
    
    \item A high-fidelity simulation framework has been constructed. Users can validate the performance of their own autonomous driving algorithms.
    \item A comparative experiment with two kinds of autonomous driving perception algorithms has been conducted to verify that our simulation method and framework can evaluate different autonomous driving algorithms with detailed and quantitative pedestrian safety indexes.

\end{enumerate}

\section{Related Work}
In this section, we systematically review the related work from three aspects: (1) Autonomous driving simulation framework and its key components;  (2) Critical characteristics of pedestrians; (3) Evaluation of pedestrian safety.

\subsection{Autonomous Driving Simulation framework}

From previous research~\cite{mukherjee2020comprehensive,zhu2021roles}, there are three key factors that distinctly influence pedestrian safety: environmental conditions, traffic conditions, and pedestrian characteristics. 

The environmental conditions are critical for constructing a high-fidelity simulation framework. First of all, the geometric designs of the road, such as the number of traffic lanes, crosswalk length~\cite{supernak2013pedestrian,diependaele2019non}, presence of central refuge and guardrail ~\cite{xu2013pedestrians}, have influential effects on pedestrian compliance to pedestrian signals, which greatly impact the safety of the pedestrian. Besides, some factors have complex effects on pedestrian safety, such as urban street trees, which can be considered roadside hazards as they were considered to reduce the visibility on roadways~\cite{budzynski2016trees} and blends the pedestrians. Thus, this paper constructs a digital twin for the target road, which maintains the environmental condition of the roads as much as possible.

For the traffic conditions, Sun et al.~\cite{sun2015estimation} found that the speed of approaching vehicles affects the pedestrians underestimating the vehicle speed, which increases the collision risk. Koh et al.~\cite{koh2014safety} presented that the behavior of pedestrians also depends on the volume of conflicting vehicle streams. For instance, increases in the volume of conflicting vehicle streams and the speed of approaching vehicles are correlated with the reduction of the red light running rate.

From the previous research on pedestrian safety factors, we can see that environmental conditions and traffic conditions are critical for pedestrian safety-related simulation, which inspired our simulation framework. Moreover, the pedestrian simulation will be described in the next subsection.

\subsection{Critical characteristics of pedestrians}
The pedestrian characteristics, such as age and gender, have a significant effect on pedestrian behavior and pedestrian safety. The majority of studies indicated that the propensity of red light running of male pedestrians was higher than that of females\cite{xie2017pedestrian,zhu2021roles}. Wang et al.~\cite{wang2018analysis} analyzed the factors of road traffic injuries in China. They found that the death/injury rate is higher in males than females, there are more death in older pedestrians and more injuries among young and middle-aged pedestrians. It is attributed to the reduction of locomotion and degradation of perception and cognitive skills by age\cite{dommes2013functional}.

From a macro point of view, the demographic characteristics of pedestrians in a specific area will affect the incidence of safety-critical events of pedestrians, so the demographic characteristics of pedestrians should be fully considered in the evaluation of pedestrian safety ~\cite{HORBERRY2019515}.

\subsection{Safety evaluation of pedestrian}

Together with the development of the traffic system and pedestrian protection technology, the evaluation of pedestrian safety has been long discussed in real-world scenarios. One of the most commonly used metrics was the collision rate \cite{ kang2019identifying}, which statistically revealed the existence of a hazard. However, a single measure of absolute collision probability can ignore in-depth and underlying information, such as pedestrian injury severity and pre-collision exposure.

Multiple features of both humans and vehicles have been considered as impact factors on the severity of pedestrian injuries, such as speed, age, impact body part, etc \cite{zajac2003factors, pour2016investigating}.
Models have been developed to estimate the relation and sensitivity between specific factors and collision risk. 
Collision velocity and human age have been proven to be critical factors for the estimation\cite{cuerden2007pedestrians, kong2010logistic, abay2013examining}.
Washington et al. \cite{washington2011statistical} offered an S-shape risk curve, which was a function of impact speed, to evaluate the injury probability by implementing logistic regressions. This model has been further expanded to consider pedestrians' age as a variable \cite{saade2020pedestrian}.
 
As collision events are relatively infrequent and have a short effective observation period, traffic conflicts became a supplementary quantitative analysis before a potential hazard \cite{songchitruksa2006extreme}. Surrogate safety measures (SSM) have been developed as a tool to assess conflicts by applying non-collision data \cite{hayward1971near}. Indicators including time \cite{chen2017surrogate}, distance \cite{golakiya2020evaluating}, and speed \cite{svensson1998method} were regarded as the most significant factors when considering pedestrian-vehicle interaction. Amini et al. \cite{amini2022development} proposed an SSM method combining three indicators mentioned above, which were minimum future relative distance (MD), time to minimum distance (TMD), and conflicting speed (CS). The result of the F-score validated this model as a good conflict classifier.

Although the safety of pedestrians has been studied in real-world scenarios, there is a gap in the evaluation of pedestrian safety in autonomous driving scenarios, which is of great importance for the industrialization of autonomous driving.

\begin{figure*}
\centering
  \includegraphics[width=1.6\columnwidth]{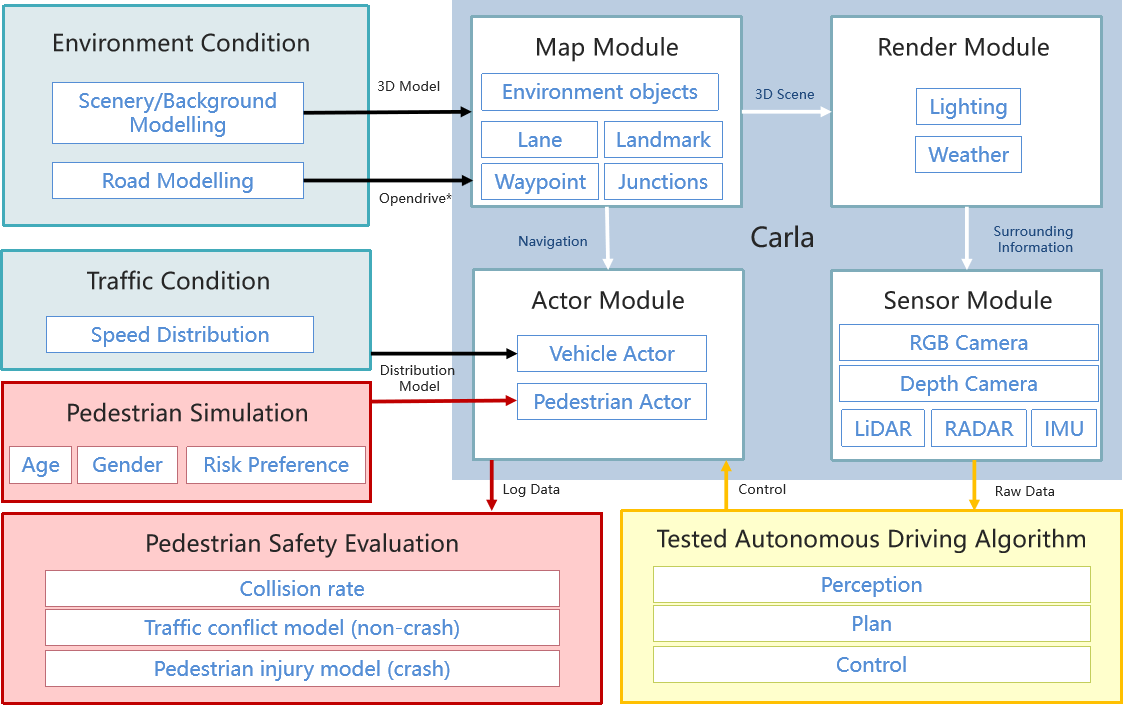}
  \caption{The proposed high fidelity simulation framework based on CARLA.}
  \label{fig:framework}
\end{figure*}

\section{High-fidelity simulation environment framework}
As shown in Fig.~\ref{fig:framework}, the proposed simulation framework is built based on CARLA~\cite{Dosovitskiy17}, consisting of five key components: environment condition, traffic condition, pedestrian model, tested autonomous driving algorithm, and safety evaluation. Researchers could replace the tested autonomous driving algorithm with their self-defined perception, plan, and control module, thus we just leave it out in this part. This section describes the environmental conditions and traffic conditions in detail. The pedestrian simulation and pedestrian safety evaluation will be presented in the next section. 

\subsection{Environment Condition Reconstruction}
In order to support simulation in a digital twin of a realistic environment, our framework supports realistic and complex road network importing, including intersections, bridges, roundabouts, turnouts, and road markings. Both the road information (including lanes, landmarks, junctions, and waypoints) and environment objects (including buildings, trees, traffic signs, etc.) can be reconstructed from a real road environment.

The road information based on a real-world map is created through OpenStreetMap\footnote{\url{https://www.openstreetmap.org/}} and Roadrunner\footnote{\url{https://www.mathworks.com/products/roadrunner.html}}. After downloading the main road topology information from OpenStreetMap, the framework uses the Roadrunner to add a lane, lane type, and landmark details.
Besides, the buildings, trees, and traffic signs are modeled from pictures in Blender and then imported to CARLA. As shown in Fig.~\ref{fig:digitaltwin}. 




\subsection{Traffic Condition Reconstruction}

In order to simulate real traffic scenarios more realistically in the simulated environment, the digital data is set referring to the traffic data distribution in the real world, including the non-intersection situation and the intersection situation, which is deployed in the form of fitted distributions. 

Studies based on real traffic data show that vehicle speed distribution follows a log-normal distribution, and vehicle headway distribution in the same lane follows a negative exponential distribution \cite{ZHOU2022103909}. In this paper, vehicle position distribution in different lanes is considered to be independent. Using the maximum likelihood equivalence estimation method, we make statistical analysis on nearly 30,000 real-world data, including intersection and non-intersection, and obtain the distribution parameters by fitting and then get the distribution function of several key scenarios.

The negative exponential distribution of headway distribution in the same lane is:
$$ P(h) = 0.1742e^{-0.1742h} \eqno{(1)} $$
The log-normal distribution of speed distribution in non-intersection is:
$$ P_{non-intersection}(v) = \frac{1}{0.4857v\sqrt{2\pi}}e^{\frac{(\ln{v}-1.8304)^2}{0.4718}}, v>0 \eqno{(2)}$$
The log-normal distribution of speed distribution in an intersection is:
$$ P_{intersection}(v) = \frac{1}{0.3827v\sqrt{2\pi}}e^{\frac{(\ln{v}-1.5853)^2}{1.5269}}, v>0 \eqno{(3)}$$

The vehicles are generated randomly according to the fitting distribution. In this way, the key features (headway and speed) of the vehicles in the simulation environment are consistent with those in the real environment.


\section{Evaluation method for pedestrian safety}
As mentioned above, pedestrian characteristics are a key factor in a high-fidelity simulation environment framework. Accurate pedestrian characteristic modeling facilitates the fidelity of the simulation environment and then improves the accuracy of pedestrian safety evaluation. First, this part focuses on safety-critical pedestrian characteristics, and how to derive them from historical data and then integrate them into a simulation environment. Then, on the basis of the pedestrian feature embedded into the framework, the safety of pedestrians is modeled, and the quantitative results are obtained.

\subsection{Pedestrian safety-critical characteristics embedding}
The main factors affecting pedestrian speed are age, gender, and risk preference are the key factors related to speed and safety~\cite{fitzpatrick2006another}. 

The distribution of pedestrians' speed varies greatly among people with different characteristics. The empirical distributions of pedestrians' speed are greatly heterogeneous with respect to pedestrians' ages. The average walking speed is 4.46m/s for teenagers aged 13-18, 1.46m/s for the young group aged 19 to 30,  1.45m/s for the middle-aged edged 31-36, and 1.03m/s for the older group aged more than 60. Limited by the sample size, the age group under 12 was eliminated, which should be considered when doing the pedestrian modeling. Analogously, the empirical distributions of pedestrians' speeds are also heterogeneous with respect to gender. 

According to our observation, the risk preference is reflected in pedestrians' behavior when faced with an approaching vehicle, which is categorized into three types: (1) pedestrians are not aware of the risk, so he/she would maintain the speed. (2) Pedestrians are aware of the risk, and he/she would speed up to pass first. (3) Pedestrians are aware of the risk, and they would step backward to yield. 

Considering the heterogeneity of speed distribution, the speeds are modeled after they are grouped by these characteristics with different proportions as weight.

So far, we have completed the speed modeling conditional on different characteristics and constructed the mixed empirical distribution model of the whole pedestrian population. Then we can generate pedestrians and their characteristics randomly according to the empirical distribution, and integrate pedestrians with different characteristics into CARLA, see the red area marked as a pedestrian simulation in Fig. \ref{fig:framework}. In this way, the pedestrian characteristics in the simulation environment are the same as the empirical distribution calculated in the real environment.

\subsection{Estimation Model of Safety}
In this part, we evaluate the safety of pedestrians from three perspectives, see the red area labeled as pedestrian evaluation in Fig \ref{fig:framework}. Firstly, collision rate, a common evaluation index of pedestrian safety, is introduced. Then, we introduce the pedestrian injury severity model under the condition of collision. Finally, we introduce the traffic conflict model to measure pedestrian safety in the case of no collision but conflict. We first distinguish collision and conflict according to ~\cite{amini2022development}.The speed-depended evidence for conflicts was offered by a indicator CS proposed in \cite{schmidt2019hacking} higher than 1m/s.
\paragraph{Collision rate} With the established simulation framework, we can test user-defined autonomous driving algorithms modules such as collaborative perception, plan, and decision. In this way, we directly obtain the collision rate from the CARLA's simulated log data.
\paragraph{Injury severity} Similar to the intensity of collision between vehicles, injury severity is used to measure the intensity of collision between vehicle and pedestrian, indicating the severity of pedestrian injuries after the occurrence of collision events. Logit$(V^2+A)$ has been proved to have the best performance in estimating the injury rate\cite{saade2020pedestrian}, thus the pedestrian injury severity can be estimated by fitting the following multivariate logistic regression model.
$$
P_{I}= \frac{p_{collision}*e^{-2.9893+0.0013*V^2+0.0286*A}}{1+e^{-2.9893+0.0013*V^2+0.0286*A}} \eqno{(4)} \label{equ:sf}
$$
where $V$ is the velocity of the vehicle and $A$ is the age of the pedestrian.
\paragraph{Conflict} As mentioned above, even if no collision happens, threats exist to the safety of pedestrians. Therefore, we have evaluated the safety in the case of no collision but conflict. Here, we use the distance between the vehicle and the pedestrian when the vehicle senses the pedestrian. The larger the distance, the sooner the vehicle senses the pedestrian and makes a timely response (such as slowing down), so it is much safer for pedestrians and vice versa.




\section{Evaluation}
We demonstrate our simulation framework and pedestrian safety evaluation with a comparative experiment with two kinds of autonomous driving perception algorithms—single-vehicle perception and V2I cooperative perception. That is the yellow area labeled as a tested autonomous driving algorithm in Fig. \ref{fig:framework} is implemented by both single-vehicle perception and V2I cooperative perception algorithm to verify their safety performance. 
\subsection{Scenario Set-up}
\begin{figure}[htp]
    \centering
    \includegraphics[scale=0.6]{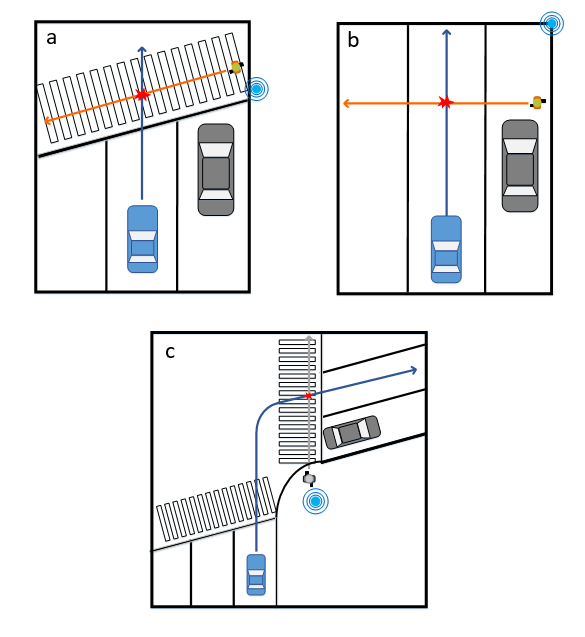}
    \caption{Schematic diagram of Scenarios. a.Crossing. b.Jaywalking. c.Background-blending. The blue circle represents the roadside camera.}
    \label{fig:diagram}
\end{figure}

To improve the test efficiency, many approaches test the autonomous driving algorithm in purposely generated scenarios that are more safety-critical. 
The chosen crowded urban intersection has a high risk for pedestrians, within which we extract three of the most safety-critical scenarios according to experience and inspection. We describe the three scenarios in detail:
\begin{itemize}
    \item Crossing: A pedestrian is crossing the road. The target blue vehicle cannot timely perceive the presence of pedestrians and recognize the intention of pedestrians due to the blocking of the black vehicle. Consequently, the target vehicle cannot timely slow down and then put pedestrians in danger. Seeing Fig.~\ref{fig:diagram}.a.
    \item Jaywalking: A pedestrian is Jaywalking. Similar to the first scenario, the pedestrian is in the blind spot of the blue vehicle due to the blocking of the black vehicles. This scenario is much more dangerous than the first scenario because the vehicle slows down before the sidewalk, but the speed remains stable during the normal driving road. Seeing Fig.~\ref{fig:diagram}.b.
    \item Background-blending: A pedestrian is indistinguishable from his/her background. On the one hand, the color of the pedestrian's clothing is the same as that of the vehicle in the background, so the camera cannot mark the pedestrian. On the other hand, the distance between the pedestrian and the waiting vehicle in the background is relatively close, so it cannot be identified by LIDAR. Seeing Fig.~\ref{fig:diagram}.c.
    
\end{itemize}

As shown in Fig. \ref{fig:diagram}, pedestrian perception equipment includes two aspects: the onboard perception equipment on the tested blue vehicle and the roadside perception equipment represented by the blue radar token. The two perception algorithms are as follows: (1) Single-vehicle perception algorithm, in which only the onboard perception equipment is used. (2) V2I collaborative algorithm, in which the perception facilities from both the vehicle and roadside are used. In V2I collaborative perception, the location of pedestrians is transformed among three coordinate systems: roadside camera, on-board camera, and real-world coordinate system, with the intrinsic matrix as follows:
$$
P_V=R_V R_I^{-1} P_I  \eqno{(5)}
$$
$$
\left(\begin{array}{l}
u_V \\
v_V \\
1
\end{array}\right)=\frac{1}{Z_V} \cdot\left(\begin{array}{ccc}
f_x & 0 & c_x \\
0 & f_y & c_y \\
0 & 0 & 1
\end{array}\right) \cdot\left(\begin{array}{l}
X_V \\
Y_V \\
Z_V
\end{array}\right) \triangleq \frac{1}{Z_V} K P_V  \eqno{(6)}
$$
where $P_V=\left(X_V, Y_V, Z_V\right)^{\top}$ and $P_I$ are respectively the positions in the on-board camera and the roadside camera, $R_{I}$ and $R_{V}$ are respectively the transformation matrices from the world to the roadside camera and the vehicle camera coordinate system; $\left(u_V, v_V\right)^{\top}$ is the position of objects in the pixel plane on the on-board camera. $f_x=f_y=  width/(2\tan\left(\operatorname{rad}\left(fov\right)\right)$, $c_x=width/2$, $c_y=height/2$, where $width$ and $height$ are the width and height of the image in the on-board camera pixel plane; $fov$ is the  horizontal Field Of View in degrees.

\subsection{Experiment details}
A $Scenario$ is defined as a specific combination of vehicles, pedestrians, and their behavior in a certain environment. The traffic participants will go to the target location according to the traffic rules.

When conducting the experiment, a controller will first load map information and switch perspective to an appropriate position to observe the simulation. Then, the controller will randomly generate roles according to the construction of traffic conditions and pedestrian simulation module, sequentially mount a preset perception and control model for each role, and specify the target location for each role, then the experiment starts, and the controller will start recording the experiment. When all the characters arrive at the target location or the experiment exceeds the preset time, the controller will clear all characters in the scenario and conduct the next experiment.

Each experiment will produce a result file recording collision information, and a global information file; the collision information file can be used for basic data analysis, while the global information file can be used for case search and culling, as well as for further data analysis.



\subsection{Overall evaluation of pedestrian safety}
\begin{table}[h]
\caption{Results of experiment in three scenarios.}
\label{results}
\begin{center}
\begin{tabular}{ccccc}
\hline
Scenario                       & Condition           & Collision & Injury   & Conflict\\ \hline

\multirow{2}{*}{Crossing}      & SV      & 0.23      & 0.0506   & 0.25 \\ \cline{2-5} 
                               & V2I                 & 0.0075    & 0.00195  & 0.05\\ \hline
\multirow{2}{*}{Jaywalking}    & SV      & 0.38      & 0.11     & 0.28 \\ \cline{2-5} 
                               & V2I                 & 0.0079    & 0.0028   & 0.04  \\ \hline
\multirow{2}{*}{Background-blending} & SV      & 0.062     & 0.0017   & 0.31\\ \cline{2-5} 
                               & V2I                 & 0         & 0        & 0.003\\ \hline
\end{tabular}
\end{center}
\end{table}
According to the above experimental settings, experiments were carried out on the three corner cases, and each scenario was tested more than 10000 times, see TABLE~\ref{results} for the overall experimental results.  The V2I collaborative perception method and the benchmark, single vehicle perception method, are embedded into the simulation framework as test algorithms, respectively. The results show that Jaywalking scenario includes a far greater threat to pedestrian safety than the other two scenarios, but with the addition of V2I cooperative perception,  its security level achieves nearly the same as crossing scenarios.  This means that when pedestrians don't obey traffic rules, V2I with cooperative perception remarkably enhanced the safety of autonomous vehicles and pedestrians. The Background-blending scenario is the safest for pedestrians, because RGB cameras and LIDAR complement each other, and the probability of perception failure is small.

\subsection{Evaluation of pedestrian injury severity}

\begin{figure}
\centering
  \includegraphics[width=0.9\columnwidth]{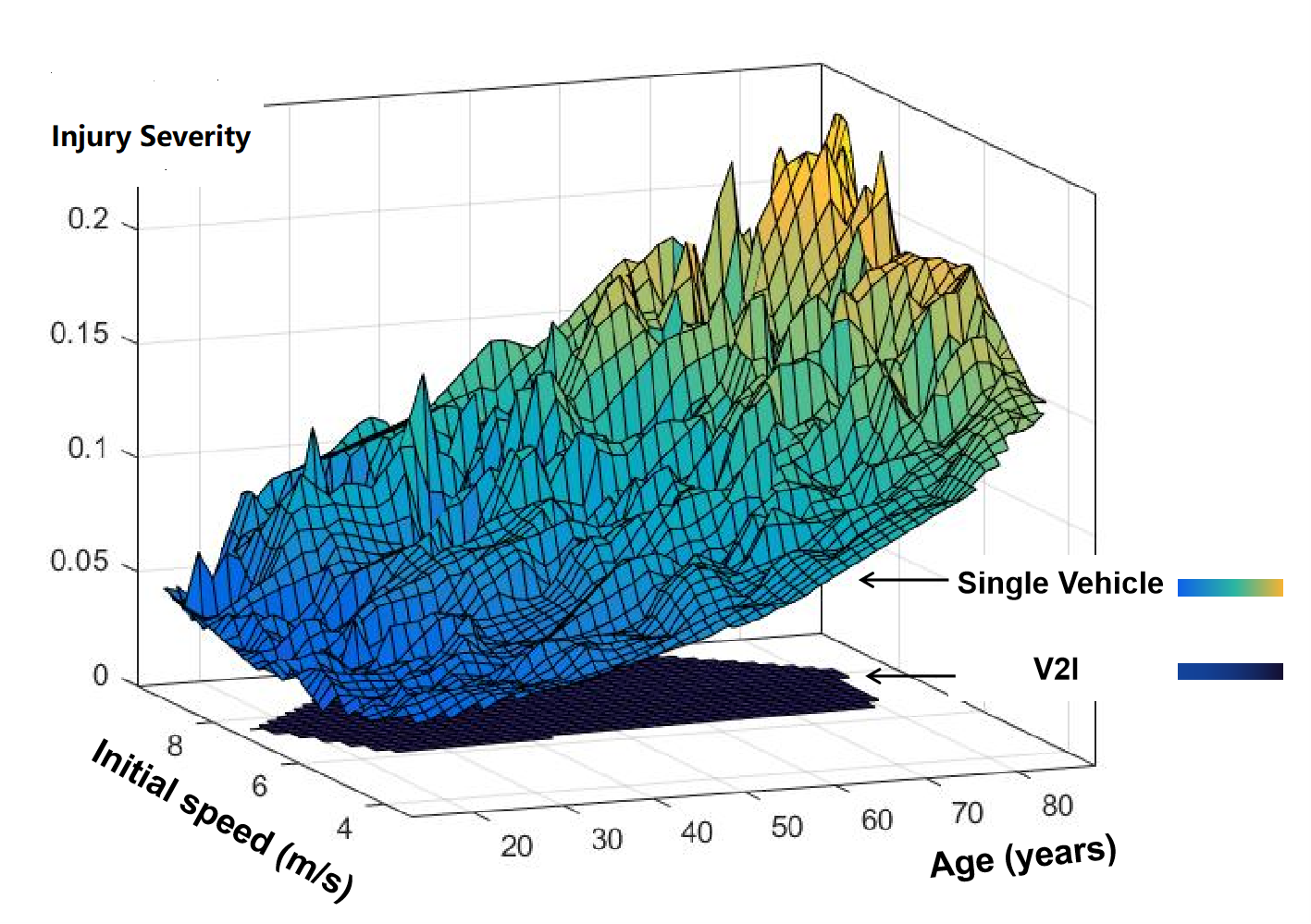}
  \caption{The injury severity curve of single vehicle and V2I condition in the Jaywalking scenario}
  \label{fig:scurve}
\end{figure}

This section conducts an in-depth study on pedestrian safety by analyzing the injury severity of pedestrians in the scenario with the lowest pedestrian safety, the Jatwalking scenario. The results are shown in Fig. \ref{fig:scurve}.

In the case of the single-vehicle perception method, the pedestrian injury severity increases as age increases, which indicates that the age structure of pedestrians under current road conditions should be fully considered when evaluating the safety of pedestrians. With the increase in initial vehicle speed, the pedestrian injury severity also increases significantly. The pedestrian injury severity is significantly lower than that of the single-vehicle perception algorithm. Therefore, V2I collaborative perception method is an effective scheme to improve pedestrian safety. This is consistent with common sense and further illustrates the validity of the proposed simulation framework.

\subsection{Evaluation of pedestrian safety in conflict condition }
\begin{figure}
\centering
  \includegraphics[width=0.9\columnwidth]{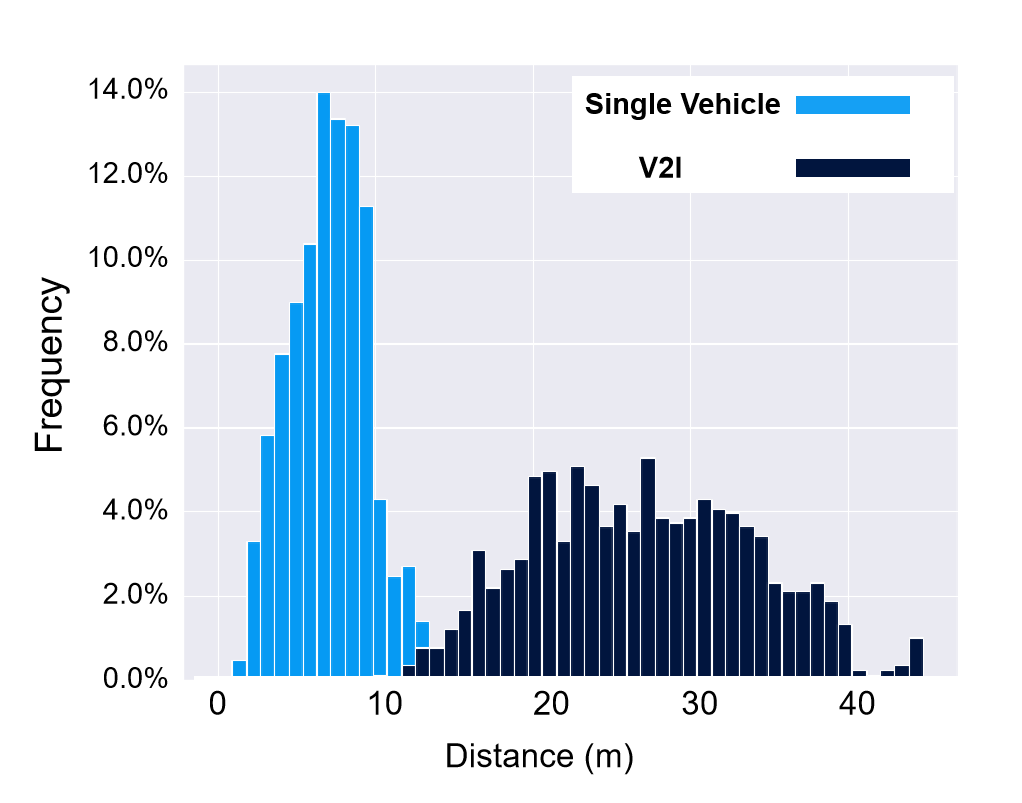}
  \caption{Human-vehicle distance distribution when the sensor module detected the pedestrian for the first time.}
  \label{fig:distribution}
\end{figure}

\begin{figure}
    \centering
    \includegraphics[width=\columnwidth]{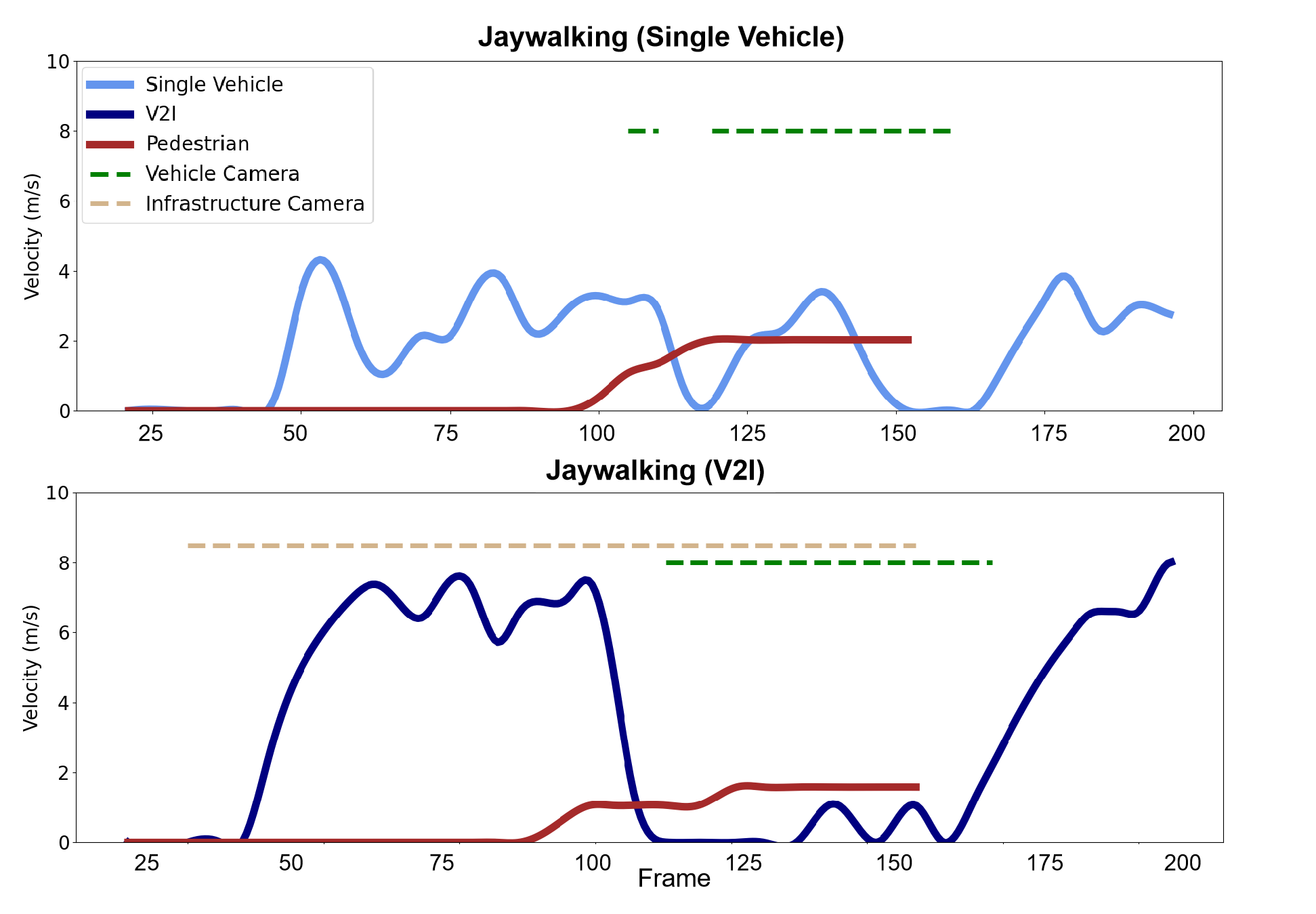}
    \caption{A velocity-frame case visualization in the jaywalking scenario. The dashed line represents the on-board and roadside detection for pedestrians.}
    \label{fig:case}
\end{figure}
This section measures pedestrian safety in the scenario where no collision but conflict occurs. The results are shown in Fig.~\ref{fig:distribution}, where the bar chart represents the distance between the car and the pedestrian when the vehicle first perceives the pedestrian. The results show that V2I is able to perceive pedestrians at greater distances.
With the V2I cooperative perception method, vehicles can detect pedestrians at a distance of more than 13 meters in 90\% of the episodes.
With only the single vehicle perception, the vehicle needs to approach the location that is within 13 meters from the pedestrians to detect the pedestrians in 90\% episodes. Therefore, with the V2I perception method, there is sufficient time to plan in advance, so the safety of pedestrians is higher.

Fig.~\ref{fig:case} shows one episode of Jaywalking scenario.   The initial states are the same.   With the advance of time, vehicle perceives pedestrians at the thirty-fifth frame under V2I collaborative perception, which ensures enough time to plan in advance.   The vehicle keeps running at a stable speed before approaching pedestrians, and not slows down until a certain distance from pedestrians.   However, the single-vehicle does not recognize the pedestrian until the one-hundredth frame 100 and then slows down sharply when it was near the pedestrian.   Therefore, V2I collaborative perception is much beneficial to pedestrian safety.



\section{CONCLUSIONS}

In this paper, we propose an evaluation method of pedestrian safety and a new high-fidelity simulation framework for autonomous driving. We comprehensively consider not only the collision events but also the conflict events together with the characteristics of pedestrians, which construct a comprehensive pedestrian safety evaluation system.  Moreover, we propose a CARLA-based high-fidelity simulation framework to validate the evaluation of pedestrian safety. We extract the characteristic of pedestrians from a real-world open data set to establish the pedestrian simulation module. Users can embed self-defined perception, plan, and control algorithms into the proposed framework and validate the performance of their algorithms on pedestrian safety. Finally, we take V2I collaborative perception as an example module to verify the validity and rationality of the proposed pedestrian safety evaluation on this simulation framework.

\bibliographystyle{IEEEtran}
\bibliography{IEEEfull,root}

\begin{thebibliography}{10}
\providecommand{\url}[1]{#1}
\csname url@rmstyle\endcsname
\providecommand{\newblock}{\relax}
\providecommand{\bibinfo}[2]{#2}
\providecommand\BIBentrySTDinterwordspacing{\spaceskip=0pt\relax}
\providecommand\BIBentryALTinterwordstretchfactor{4}
\providecommand\BIBentryALTinterwordspacing{\spaceskip=\fontdimen2\font plus
\BIBentryALTinterwordstretchfactor\fontdimen3\font minus
  \fontdimen4\font\relax}
\providecommand\BIBforeignlanguage[2]{{%
\expandafter\ifx\csname l@#1\endcsname\relax
\typeout{** WARNING: IEEEtran.bst: No hyphenation pattern has been}%
\typeout{** loaded for the language `#1'. Using the pattern for}%
\typeout{** the default language instead.}%
\else
\language=\csname l@#1\endcsname
\fi
#2}}

\bibitem{world2015global}
W.~H. Organization, \emph{Global status report on road safety 2015}.\hskip 1em
  plus 0.5em minus 0.4em\relax World Health Organization, 2015.

\bibitem{SHEYKHFARD2021681}
A.~Sheykhfard, F.~Haghighi, E.~Papadimitriou, and P.~{Van Gelder}, ``Review and
  assessment of different perspectives of vehicle-pedestrian conflicts and
  crashes: Passive and active analysis approaches,'' \emph{Journal of Traffic
  and Transportation Engineering (English Edition)}, vol.~8, no.~5, pp.
  681--702, 2021.

\bibitem{su15010858}
S.~Gong, S.~Feng, L.~Yu, S.~Wang, Y.~Chen, Q.~Zhang, Q.~Xu, and T.~Zhou, ``A
  study on the psychological field model of drivers in traffic conflict
  environments,'' \emph{Sustainability}, vol.~15, no.~1, 2023.

\bibitem{LIU2021115}
Y.~Liu, R.~Alsaleh, and T.~Sayed, ``Modeling the influence of mobile phone use
  distraction on pedestrian reaction times to green signals: A multilevel
  mixed-effects parametric survival model,'' \emph{Transportation Research Part
  F: Traffic Psychology and Behaviour}, vol.~81, pp. 115--129, 2021.

\bibitem{DELEN2006434}
D.~Delen, R.~Sharda, and M.~Bessonov, ``Identifying significant predictors of
  injury severity in traffic accidents using a series of artificial neural
  networks,'' \emph{Accident Analysis \& Prevention}, vol.~38, no.~3, pp.
  434--444, 2006.

\bibitem{ta2020secure}
V.-T. Ta and A.~Dvir, ``A secure road traffic congestion detection and
  notification concept based on v2i communications,'' \emph{Vehicular
  Communications}, vol.~25, p. 100283, 2020.

\bibitem{jia2016enhanced}
D.~Jia and D.~Ngoduy, ``Enhanced cooperative car-following traffic model with
  the combination of v2v and v2i communication,'' \emph{Transportation Research
  Part B: Methodological}, vol.~90, pp. 172--191, 2016.

\bibitem{mughal2020cooperative}
U.~A. Mughal, J.~Xiao, I.~Ahmad, and K.~Chang, ``Cooperative resource
  management for c-v2i communications in a dense urban environment,''
  \emph{Vehicular Communications}, vol.~26, p. 100282, 2020.

\bibitem{he2021improved}
H.~He, Y.~Wang, R.~Han, M.~Han, Y.~Bai, and Q.~Liu, ``An improved mpc-based
  energy management strategy for hybrid vehicles using v2v and v2i
  communications,'' \emph{Energy}, vol. 225, p. 120273, 2021.

\bibitem{yan2018coordinated}
S.~Yan, J.-s. Wang, and J.-l. Wang, ``Coordinated control of vehicle lane
  change and speed at intersection under v2x,'' in \emph{2018 3rd International
  Conference on Mechanical, Control and Computer Engineering (ICMCCE)}.\hskip
  1em plus 0.5em minus 0.4em\relax IEEE, 2018, pp. 69--73.

\bibitem{kalra2016driving}
N.~Kalra and S.~M. Paddock, ``Driving to safety: How many miles of driving
  would it take to demonstrate autonomous vehicle reliability?''
  \emph{Transportation Research Part A: Policy and Practice}, vol.~94, pp.
  182--193, 2016.

\bibitem{mukherjee2020comprehensive}
D.~Mukherjee and S.~Mitra, ``A comprehensive study on identification of risk
  factors for fatal pedestrian crashes at urban intersections in a developing
  country,'' \emph{Asian Transport Studies}, vol.~6, p. 100003, 2020.

\bibitem{zhu2021roles}
D.~Zhu, N.~Sze, and L.~Bai, ``Roles of personal and environmental factors in
  the red light running propensity of pedestrian: Case study at the urban
  crosswalks,'' \emph{Transportation research part F: traffic psychology and
  behaviour}, vol.~76, pp. 47--58, 2021.

\bibitem{supernak2013pedestrian}
J.~Supernak, V.~Verma, I.~Supernak, \emph{et~al.}, ``Pedestrian countdown
  signals: what impact on safe crossing?'' \emph{Open journal of civil
  Engineering}, vol.~3, no.~03, p.~39, 2013.

\bibitem{diependaele2019non}
K.~Diependaele, ``Non-compliance with pedestrian traffic lights in belgian
  cities,'' \emph{Transportation Research Part F: Traffic Psychology and
  Behaviour}, vol.~67, pp. 230--241, 2019.

\bibitem{xu2013pedestrians}
Y.~Xu, Y.~Li, and F.~Zhang, ``Pedestrians’ intention to jaywalk: Automatic or
  planned? a study based on a dual-process model in china,'' \emph{Accident
  Analysis \& Prevention}, vol.~50, pp. 811--819, 2013.

\bibitem{budzynski2016trees}
M.~Budzynski, K.~Jamroz, L.~Jelinski, and M.~Antoniuk, ``Why are trees still
  such a major hazard to drivers in poland?'' \emph{Transportation research
  procedia}, vol.~14, pp. 4150--4159, 2016.

\bibitem{sun2015estimation}
R.~Sun, X.~Zhuang, C.~Wu, G.~Zhao, and K.~Zhang, ``The estimation of vehicle
  speed and stopping distance by pedestrians crossing streets in a naturalistic
  traffic environment,'' \emph{Transportation research part F: traffic
  psychology and behaviour}, vol.~30, pp. 97--106, 2015.

\bibitem{koh2014safety}
P.~Koh, Y.~Wong, and P.~Chandrasekar, ``Safety evaluation of pedestrian
  behaviour and violations at signalised pedestrian crossings,'' \emph{Safety
  science}, vol.~70, pp. 143--152, 2014.

\bibitem{xie2017pedestrian}
S.~Xie, S.~Wong, T.~M. Ng, and W.~H. Lam, ``Pedestrian crossing behavior at
  signalized crosswalks,'' \emph{Journal of transportation engineering, Part A:
  Systems}, vol. 143, no.~8, p. 04017036, 2017.

\bibitem{wang2018analysis}
L.~Wang, C.~Yu, Y.~Zhang, L.~Luo, and G.~Zhang, ``An analysis of the
  characteristics of road traffic injuries and a prediction of fatalities in
  china from 1996 to 2015,'' \emph{Traffic injury prevention}, vol.~19, no.~7,
  pp. 749--754, 2018.

\bibitem{dommes2013functional}
A.~Dommes, V.~Cavallo, and J.~Oxley, ``Functional declines as predictors of
  risky street-crossing decisions in older pedestrians,'' \emph{Accident
  Analysis \& Prevention}, vol.~59, pp. 135--143, 2013.

\bibitem{HORBERRY2019515}
T.~Horberry, R.~Osborne, and K.~Young, ``Pedestrian smartphone distraction:
  Prevalence and potential severity,'' \emph{Transportation Research Part F:
  Traffic Psychology and Behaviour}, vol.~60, pp. 515--523, 2019.

\bibitem{kang2019identifying}
B.~Kang, ``Identifying street design elements associated with
  vehicle-to-pedestrian collision reduction at intersections in new york
  city,'' \emph{Accident Analysis \& Prevention}, vol. 122, pp. 308--317, 2019.

\bibitem{zajac2003factors}
S.~S. Zajac and J.~N. Ivan, ``Factors influencing injury severity of motor
  vehicle--crossing pedestrian crashes in rural connecticut,'' \emph{Accident
  Analysis \& Prevention}, vol.~35, no.~3, pp. 369--379, 2003.

\bibitem{pour2016investigating}
M.~Pour-Rouholamin and H.~Zhou, ``Investigating the risk factors associated
  with pedestrian injury severity in illinois,'' \emph{Journal of safety
  research}, vol.~57, pp. 9--17, 2016.

\bibitem{cuerden2007pedestrians}
R.~Cuerden, D.~Richards, and J.~Hill, ``Pedestrians and their survivability at
  different impact speeds,'' in \emph{Proceedings of the 20th International
  Technical Conference on the Enhanced Safety of Vehicles, Lyon, France,
  Paper}, no. 07-0440, 2007.

\bibitem{kong2010logistic}
C.~Kong and J.~Yang, ``Logistic regression analysis of pedestrian casualty risk
  in passenger vehicle collisions in china,'' \emph{Accident Analysis \&
  Prevention}, vol.~42, no.~4, pp. 987--993, 2010.

\bibitem{abay2013examining}
K.~A. Abay, ``Examining pedestrian-injury severity using alternative
  disaggregate models,'' \emph{Research in Transportation Economics}, vol.~43,
  no.~1, pp. 123--136, 2013.

\bibitem{washington2011statistical}
S.~Washington, M.~Karlaftis, and F.~Mannering, ``Statistical and econometric
  methods for transportation data analysis,'' 2011.

\bibitem{saade2020pedestrian}
J.~Saad{\'e}, S.~Cuny, M.~Labrousse, E.~Song, C.~Chauvel, and P.~Chr{\'e}tien,
  ``Pedestrian injuries and vehicles-related risk factors in car-to-pedestrian
  frontal collisions,'' in \emph{Proceedings of the 2020 IRCOBI Conference
  Proceedings}.\hskip 1em plus 0.5em minus 0.4em\relax IRCOBI Munich, 2020, pp.
  278--289.

\bibitem{songchitruksa2006extreme}
P.~Songchitruksa and A.~P. Tarko, ``The extreme value theory approach to safety
  estimation,'' \emph{Accident Analysis \& Prevention}, vol.~38, no.~4, pp.
  811--822, 2006.

\bibitem{hayward1971near}
J.~Hayward, \emph{Near misses as a measure of safety at urban
  intersections}.\hskip 1em plus 0.5em minus 0.4em\relax Pennsylvania
  Transportation and Traffic Safety Center, 1971.

\bibitem{chen2017surrogate}
P.~Chen, W.~Zeng, G.~Yu, and Y.~Wang, ``Surrogate safety analysis of
  pedestrian-vehicle conflict at intersections using unmanned aerial vehicle
  videos,'' \emph{Journal of advanced transportation}, vol. 2017, 2017.

\bibitem{golakiya2020evaluating}
H.~D. Golakiya, R.~Chauhan, and A.~Dhamaniya, ``Evaluating safe distance for
  pedestrians on urban midblock sections using trajectory plots,''
  \emph{European Transport$\backslash$Trasporti Europei}, vol.~75, no.~2, 2020.

\bibitem{svensson1998method}
A.~Svensson, \emph{A method for analysing the traffic process in a safety
  perspective}.\hskip 1em plus 0.5em minus 0.4em\relax Lund Institute of
  Technology Sweden, 1998.

\bibitem{amini2022development}
R.~E. Amini, K.~Yang, and C.~Antoniou, ``Development of a conflict risk
  evaluation model to assess pedestrian safety in interaction with vehicles,''
  \emph{Accident Analysis \& Prevention}, vol. 175, p. 106773, 2022.

\bibitem{Dosovitskiy17}
A.~Dosovitskiy, G.~Ros, F.~Codevilla, A.~Lopez, and V.~Koltun, ``{CARLA}: {An}
  open urban driving simulator,'' in \emph{Proceedings of the 1st Annual
  Conference on Robot Learning}, 2017, pp. 1--16.

\bibitem{ZHOU2022103909}
Z.~Zhou, Z.~Yang, Y.~Zhang, Y.~Huang, H.~Chen, and Z.~Yu, ``A comprehensive
  study of speed prediction in transportation system: From vehicle to
  traffic,'' \emph{iScience}, vol.~25, no.~3, p. 103909, 2022.

\bibitem{fitzpatrick2006another}
K.~Fitzpatrick, M.~A. Brewer, and S.~Turner, ``Another look at pedestrian
  walking speed,'' \emph{Transportation research record}, vol. 1982, no.~1, pp.
  21--29, 2006.

\bibitem{schmidt2019hacking}
H.~Schmidt, J.~Terwilliger, D.~AlAdawy, and L.~Fridman, ``Hacking nonverbal
  communication between pedestrians and vehicles in virtual reality,''
  \emph{arXiv preprint arXiv:1904.01931}, 2019.

\end{thebibliography}

\end{document}